\documentclass{article}
\usepackage[preprint]{neurips_2026}

\usepackage[utf8]{inputenc}
\usepackage[T1]{fontenc}

\usepackage{comment}

\usepackage{amsmath}
\usepackage{amssymb}
\usepackage{amsfonts}
\usepackage{mathtools}
\usepackage{nicefrac}

\usepackage[table]{xcolor}

\definecolor{bestcolor}{HTML}{D9EAD3}
\definecolor{secondcolor}{HTML}{FFF2CC}

\usepackage{booktabs}
\usepackage{multirow}

\usepackage{algorithm}
\usepackage{algorithmic}

\usepackage{graphicx}
\graphicspath{{paper_figs/PDF/}}

\usepackage{url}
\usepackage{hyperref}

\title{PILOT: Policy-Informed Learned Optimization for Adaptive Deep Network Training}

\author{
Sattam Altuuaim$^{1}$ \quad
Lama Ayash$^{1}$ \quad
Muhammad Mubashar$^{2}$ \quad
Naeemullah Khan$^{1}$ \\
\vspace{0.3cm}
$^{1}$King Abdullah University of Science and Technology (KAUST), Thuwal, Saudi Arabia \\
$^{2}$University of Strathclyde, Glasgow, Scotland \\
\texttt{\{satttam.tuuaim, lama.ayash, naeemullah.khan\}@kaust.edu.sa} \\
\texttt{muhammad.mubashar@strath.ac.uk}
}

\begin{document}

\maketitle

\begin{abstract}
Despite the central role of optimization in deep learning, most optimizers rely on update structures whose functional form is fixed before training begins. This static design can limit their ability to respond to changing gradient behavior across the loss landscape, where training may shift between stable, noisy, and inconsistent regimes.
This study proposes \textsc{PILOT} (Policy-Informed Learned OpTimizer), an online optimizer that adapts its update behavior during training. Rather than using a fixed balance between momentum, normalization, and sign-based updates, \textsc{PILOT} uses gradient-direction agreement as a signal of local training stability. Conditioning the update rule on this agreement signal allows the optimizer to adjust its behavior when gradients become stable, noisy, or inconsistent.
Experiments on FashionMNIST and CIFAR-10 show that \textsc{PILOT} consistently achieves the highest accuracy among the evaluated optimizers across convolutional settings. On the CNN architecture, \textsc{PILOT} reaches 94.13\% on FashionMNIST and 81.94\% on CIFAR-10. On ResNet-18, it further improves performance, reaching 95.71\% on FashionMNIST and 93.42\% on CIFAR-10. These results suggest that learning how to adapt the update structure during training can improve performance across both compact and deeper convolutional models while preserving a simple first-order optimization framework.
The implementation of PILOT is publicly available at
\href{https://github.com/SattamAltwaim/PILOT.git}{https://github.com/SattamAltwaim/PILOT.git}.
\end{abstract}

\section{Introduction}

\begin{figure}[t]
\centering
\includegraphics[width=0.70\linewidth]{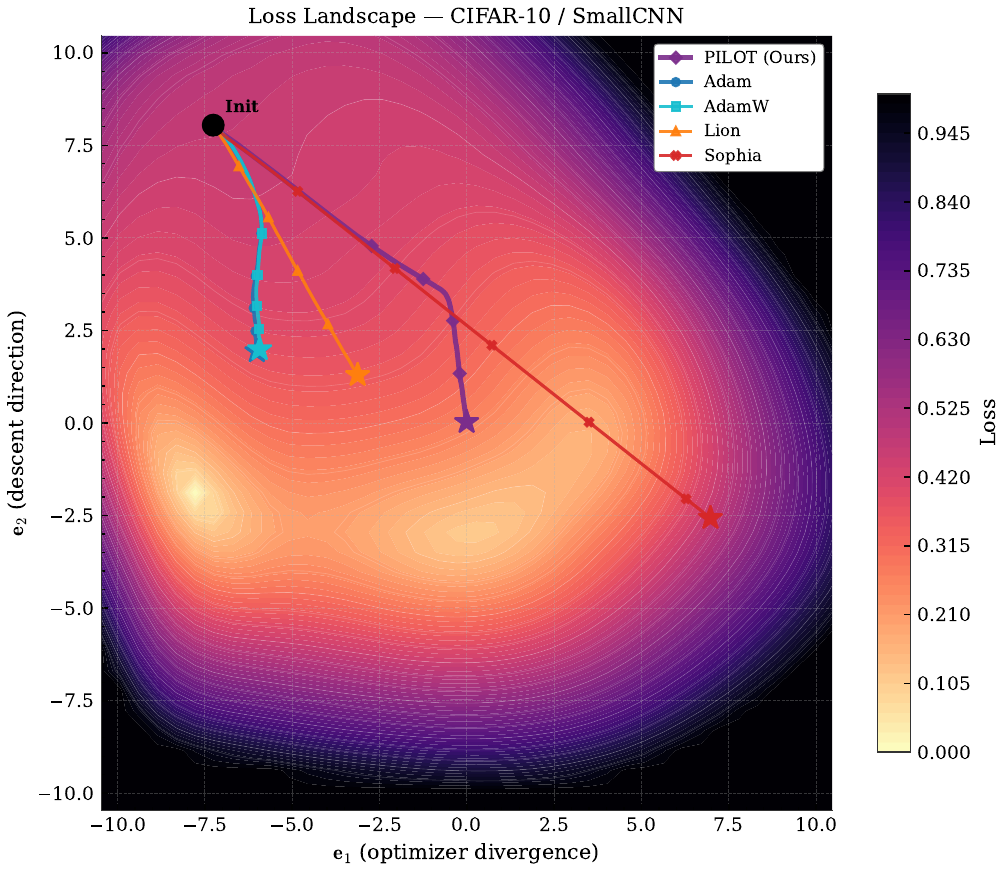}
\caption{Optimizer trajectories on the loss landscape for CIFAR-10 / SmallCNN. \textsc{PILOT} follows a distinct path through the surface and converges to a lower-loss region compared to the baselines.}
\label{fig:landscape}
\end{figure}

Optimization is a central component of deep neural network training because it determines how model parameters are updated from stochastic gradient information~\cite{belhadri2025optimizing}. The choice of optimizer directly affects convergence speed, numerical stability, generalization behavior, and computational cost, particularly as models and datasets increase in scale~\cite{altinel2025development}. Modern deep learning optimization has evolved from classical stochastic gradient descent and momentum-based methods toward adaptive, sign-based, curvature-aware, and learned update mechanisms~\cite{liu2025recent}. Recent surveys emphasize that optimizer design remains an active research area because deep learning objectives are highly non-convex, noisy, architecture-dependent, and sensitive to interactions among gradients, learning rates, curvature, and regularization~\cite{liu2025recent}. Widely used adaptive methods such as Adam~\cite{kingma2015adam} and AdamW~\cite{loshchilov2019decoupled} improve robustness through moment estimation and variance normalization, while newer optimizers such as Lion~\cite{chen2023symbolic} and Sophia~\cite{liu2023sophia} explore sign-based and curvature-informed update rules.

Despite this progress, a major limitation remains: most optimizers use a fixed update structure throughout training. Adam and AdamW impose a predefined balance between momentum and second-moment normalization through fixed moment-update equations and adaptive scaling rules~\cite{kingma2015adam,loshchilov2019decoupled}. Lion relies on sign-based updates, improving memory efficiency but discarding explicit gradient magnitude information in the parameter update~\cite{chen2023symbolic}. Sophia introduces curvature-aware scaling through Hessian-based estimation, but its update mechanism is still specified in advance and introduces additional estimation overhead~\cite{liu2023sophia}. Learned optimizers and optimizer-discovery methods expand the design space, but they often require expensive offline search, meta-training, or task-specific optimizer generation~\cite{andrychowicz2016learning,bello2017neural}. This gap motivates an online adaptive mechanism that adjusts the optimizer's update behavior during ordinary training, without requiring offline optimizer search, expensive meta-training, or task-specific optimizer generation.

The motivation behind PILOT is that the local optimization regime changes over training. Consecutive gradients may become strongly aligned in smooth regions, weakly aligned in noisy regions, or negatively aligned near unstable curvature or rapidly changing loss surfaces. Such directional agreement provides a compact signal about the current training dynamics. PILOT uses this signal to adapt the optimizer online by learning a small policy that controls three core update primitives: momentum reliance, variance-normalization strength, and sign-based behavior. Instead of selecting one fixed optimizer behavior before training, PILOT continuously interpolates between Adam-like, raw-gradient, and sign-based update regimes according to the observed gradient-agreement signal.

The main contributions are summarized as follows:
\begin{itemize}
    \item An online adaptive optimizer, PILOT, is introduced to dynamically modulate momentum interpolation, variance normalization, and sign-based behavior during training.
    \item A gradient-direction agreement signal is used as a landscape-aware indicator of local optimization stability, enabling the update rule to respond to changing training dynamics.
    \item A compact polynomial policy with only $3(d+1)$ learnable coefficients is proposed, allowing adaptive optimizer behavior without offline search or explicit second-order estimation.
    \item Existing update behaviors, including Adam-like adaptive updates and sign-based updates, are represented as fixed configurations within the same general formulation.
    \item Empirical evaluation on FashionMNIST and CIFAR-10 using CNN and ResNet-18 architectures shows that PILOT is competitive with strong optimizer baselines and can improve accuracy and stability in selected convolutional settings.
\end{itemize}

Overall, \textsc{PILOT} aims to bridge the gap between rigid hand-designed optimizers and computationally expensive learned optimizers. By learning a small online policy conditioned on gradient-direction agreement, the proposed method provides a practical mechanism for adaptive optimizer behavior while retaining a first-order optimization structure. This makes \textsc{PILOT} a promising direction for optimizer design in settings where the training landscape varies over time and a single fixed update rule may be insufficient. The qualitative loss-landscape visualization in Fig.~\ref{fig:landscape} further illustrates this behavior, showing that \textsc{PILOT} follows a distinct optimization trajectory and reaches a lower-loss region compared with the baseline optimizers.

\section{Related Work}

\subsection{Adaptive first-order methods}
Adaptive gradient-based optimizers constitute the dominant paradigm for training deep neural networks. The Adam optimizer~\cite{kingma2015adam} combines first-order momentum with element-wise second-moment normalization, yielding scale-invariant updates that improve convergence stability. AdamW~\cite{loshchilov2019decoupled} further improves generalization by decoupling weight decay from the adaptive update, addressing implicit regularization issues in Adam. Subsequent work has refined the estimation of second-order statistics; for example, AdaBelief~\cite{zhuang2020adabelief} replaces the variance of gradients with the variance of prediction error, adjusting step sizes based on the reliability of gradient estimates. While these methods achieve strong empirical performance, they share a common structural limitation: the update rule remains fixed throughout training, enforcing a static trade-off between momentum and normalization that does not adapt to changing optimization dynamics.

\subsection{Sign-based and magnitude-invariant updates}
An alternative line of work explores sign-based optimization, where update magnitudes are decoupled from gradient scale. Lion~\cite{chen2023symbolic} eliminates second-moment estimation and performs updates using the sign of a momentum term, resulting in improved memory efficiency and competitive performance in large-scale settings. Earlier approaches such as signSGD~\cite{bernstein2018signsgd} similarly demonstrate robustness under noisy gradients. However, by discarding magnitude information, these methods impose a rigid inductive bias that may be suboptimal in regimes where gradient scale encodes curvature or noise characteristics. Consequently, their ability to adapt across heterogeneous optimization regimes remains limited.

\subsection{Curvature-aware and second-order methods}
To improve convergence efficiency, several approaches incorporate curvature information into the update rule. Methods such as AdaHessian~\cite{yao2021adahessian} approximate second-order structure through stochastic Hessian estimation, while Sophia~\cite{liu2023sophia} employs a diagonal Hessian approximation with clipping to stabilize updates in large-scale language model training. These methods provide more informed scaling of parameter updates and can significantly reduce training time. However, they introduce additional computational overhead and rely on predefined mechanisms for incorporating curvature, resulting in update rules that remain structurally fixed rather than dynamically adapting to the optimization trajectory.

\subsection{Learned optimizers and automated update design}
Optimizer design has also been approached as a learning problem. Early work formulates optimizer discovery using reinforcement learning or gradient-based meta-learning, enabling the construction of parameterized update rules~\cite{andrychowicz2016learning, bello2017neural}. More recent approaches, including large-scale learned optimizers and symbolic program search methods~\cite{chen2023symbolic}, generate fixed update rules that generalize across tasks. While these methods demonstrate the potential of data-driven optimizer design, they typically require expensive offline training or search procedures and produce static update rules that are not adapted during training.

\section{Method}

This section presents the proposed \textsc{PILOT} optimizer. The formulation treats adaptive optimization as a policy-controlled update problem and uses gradient-direction agreement as an online signal of update consistency. The policy maps this signal to three variables that control momentum reliance, variance-normalization strength, and sign-based behavior. The section then defines the policy-conditioned update rule and the online policy-learning procedure.
\subsection{Problem Formulation}

Consider the optimization of a parameterized model with parameters 
\(\theta \in \mathbb{R}^D\) under a stochastic objective:
\begin{equation}
\min_{\theta} \; \mathcal{L}(\theta)
=
\mathbb{E}_{(x,y)\sim \mathcal{D}}
\left[
\ell(\theta; x, y)
\right],
\end{equation}
where \(\ell(\theta; x, y)\) denotes the sample-wise loss and 
\(\mathcal{D}\) is the data distribution. In practice, optimization is performed using stochastic gradients computed over mini-batches:
\begin{equation}
g_t = \nabla_{\theta} \mathcal{L}_t(\theta_t),
\end{equation}
where \(\mathcal{L}_t\) denotes the empirical mini-batch loss at iteration \(t\).

Many adaptive optimization methods construct parameter updates by applying a predefined transformation to the current gradient and optimizer state:
\begin{equation}
\theta_{t+1}
=
\theta_t
-
\eta \, \mathcal{U}(g_t; s_t, \psi),
\end{equation}
where \(\eta\) is the learning rate, \(s_t\) denotes time-varying optimizer states such as momentum and variance estimates, and \(\psi\) denotes fixed optimizer hyperparameters. Although the internal states evolve during training, the functional form of \(\mathcal{U}\) is usually specified a priori and remains fixed throughout optimization.

However, the optimization trajectory may encounter regions with different smoothness, curvature, and stochastic-noise characteristics. This suggests that a single fixed update structure may be suboptimal across all stages of training. The objective is therefore to construct an update mechanism in which the transformation \(\mathcal{U}\) can adapt online in response to observed optimization dynamics, rather than relying on a fixed functional form throughout training.

\subsection{PILOT Optimizer}

\textsc{PILOT} is an adaptive first-order optimizer that modulates its update behavior during training using a policy conditioned on gradient-direction agreement. The policy outputs three scalar control variables corresponding to momentum reliance, variance-normalization strength, and sign-based behavior. These variables determine how the optimizer combines momentum-based information, second-moment normalization, and sign-dominated updates during optimization.

\subsubsection{Gradient-Direction Agreement}

The first step in \textsc{PILOT} is to compute a scalar signal that summarizes the consistency of recent stochastic gradients. Consecutive gradients may become strongly aligned in smooth regions of the optimization trajectory, weakly aligned under stochastic noise, or negatively aligned when the update direction changes rapidly. Therefore, gradient-direction agreement provides a simple proxy for local update consistency.

Let \(g_t\) and \(g_{t-1}\) denote the flattened stochastic gradients at two successive iterations. For \(t \geq 2\), their directional agreement is defined using a numerically stabilized cosine similarity:
\begin{equation}
r_t =
\frac{g_t^\top g_{t-1}}
{\|g_t\|_2 \, \|g_{t-1}\|_2 + \epsilon},
\end{equation}
where \(\epsilon > 0\) is used for numerical stability. At the first iteration, \(r_1\) is initialized to zero.

The scalar \(r_t\) is bounded in \([-1,1]\). Positive values indicate that consecutive gradients point in similar directions, suggesting consistent local descent behavior. Values near zero indicate weak directional consistency, which may arise from stochastic mini-batch variation or rapidly changing gradients. Negative values indicate directional disagreement between successive updates.

To reduce short-term fluctuation, \textsc{PILOT} maintains an exponential moving average of the agreement signal:
\begin{equation}
\rho_t =
\gamma \rho_{t-1}
+
(1-\gamma) r_t,
\end{equation}
where \(\gamma \in [0,1)\) controls the smoothing strength and \(\rho_0\) is initialized to zero.

The smoothed signal \(\rho_t\) is used as the policy input for \textsc{PILOT}. Rather than treating gradient agreement as a direct estimate of curvature or smoothness, \(\rho_t\) serves as a compact online descriptor of update consistency.

\subsubsection{Learnable Policy}

To enable adaptive optimization behavior, \textsc{PILOT} uses a policy that maps the smoothed gradient-agreement signal \(\rho_t\) to scalar control variables. Specifically, the policy outputs
\begin{equation}
P_t = [p_{m,t}, p_{v,t}, p_{s,t}],
\end{equation}
where \(p_{m,t}\), \(p_{v,t}\), and \(p_{s,t}\) control momentum reliance, variance-normalization strength, and sign-based behavior, respectively.

The policy variables are parameterized as polynomial functions of \(\rho_t\), followed by sigmoid activations:
\begin{equation}
p_{m,t} = \sigma\!\bigl(f(\rho_t;\phi_m)\bigr), 
\quad
p_{s,t} = \sigma\!\bigl(f(\rho_t;\phi_s)\bigr),
\end{equation}
\begin{equation}
p_{v,t} = \tfrac{1}{2}\,\sigma\!\bigl(f(\rho_t;\phi_v)\bigr),
\end{equation}
where \(\sigma(\cdot)\) denotes the sigmoid function and each \(f(\rho;\phi_\cdot)\) is a univariate polynomial of degree \(d\):
\begin{equation}
f(\rho;\,c_0,\dots,c_d) = \sum_{k=0}^{d} c_k \rho^k .
\end{equation}
The polynomial is evaluated using Horner's method for numerical stability. The full set of learnable coefficients is
\begin{equation}
\phi = \{\phi_m,\phi_v,\phi_s\} \in \mathbb{R}^{3(d+1)}.
\end{equation}
The degree \(d\) is a hyperparameter controlling the expressiveness of the policy: \(d=1\) yields a linear response, while higher degrees allow more flexible responses to changes in gradient agreement.

Each control variable is constrained to a specific range. The parameter \(p_{m,t}\in[0,1]\) controls the interpolation between the momentum estimate and the current gradient. The parameter \(p_{v,t}\in[0,0.5]\) determines the strength of second-moment normalization, where \(p_{v,t}=0\) corresponds to no variance normalization and \(p_{v,t}=0.5\) corresponds to Adam-style square-root scaling. The parameter \(p_{s,t}\in[0,1]\) governs the degree of sign-based behavior, with larger values producing a more sign-dominated update direction.

This formulation allows several first-order update behaviors to be represented through fixed configurations of \(P_t\), while \textsc{PILOT} allows these configurations to vary dynamically during training as a function of the observed gradient-agreement signal.

\subsubsection{Policy-Controlled Update Rule}

The proposed update rule builds upon first- and second-moment estimates commonly used in adaptive optimization. The moments are computed as
\begin{equation}
m_t =
\beta_1 m_{t-1}
+
(1-\beta_1) g_t,
\quad
v_t =
\beta_2 v_{t-1}
+
(1-\beta_2)(g_t \odot g_t),
\end{equation}
with corresponding bias-corrected estimates
\begin{equation}
\hat{m}_t =
\frac{m_t}{1-\beta_1^t},
\quad
\hat{v}_t =
\frac{v_t}{1-\beta_2^t}.
\end{equation}

A policy-controlled interpolation between the momentum estimate and the current gradient is defined as
\begin{equation}
n_t =
p_{m,t}\hat{m}_t
+
(1-p_{m,t})g_t,
\end{equation}
where \(p_{m,t}\) controls momentum reliance.

The final parameter update is given by
\begin{equation}
\theta_{t+1}
=
\theta_t
-
\eta
\frac{
(|n_t|+\epsilon_n)^{\,1-p_{s,t}}
\odot
\mathrm{sign}(n_t)
}
{
\hat{v}_t^{\,p_{v,t}}+\epsilon
},
\end{equation}
where \(\eta\) is the learning rate, \(\epsilon>0\) and \(\epsilon_n>0\) are numerical-stability constants, and all powers and operations are applied element-wise. The policy variables \(p_{v,t}\) and \(p_{s,t}\) control variance-normalization strength and sign-based behavior, respectively.

This formulation connects several first-order update behaviors. Setting \(p_{m,t}=1\), \(p_{v,t}=0.5\), and \(p_{s,t}=0\) recovers an Adam-style adaptive direction without decoupled weight decay. Setting \(p_{s,t}=1\) and \(p_{v,t}=0\) yields an update proportional to \(\mathrm{sign}(n_t)\), corresponding to a sign-based direction. Intermediate values provide continuous interpolation between magnitude-sensitive, variance-normalized, and sign-dominated updates.
\subsubsection{Online Policy Learning}

The policy parameters \(\phi=\{\phi_m,\phi_v,\phi_s\}\) are updated online during training. At iteration \(t\), the policy produces the control variables \(p_{m,t}\), \(p_{v,t}\), and \(p_{s,t}\), which determine the policy-controlled parameter update. This update can be written as
\begin{equation}
\theta_{t+1}
=
\theta_t
+
\Delta \theta_t(\phi_t),
\end{equation}
where \(\Delta \theta_t(\phi_t)\) denotes the policy-controlled update increment, including the negative descent direction and learning-rate scaling defined in the update rule above.

After the subsequent mini-batch gradient
\begin{equation}
g_{t+1}
=
\nabla_{\theta}\mathcal{L}_{t+1}(\theta_{t+1})
\end{equation}
is observed, the policy is updated using a one-step meta-gradient estimate. The effect of the previous policy-controlled update on the next loss is estimated as
\begin{equation}
\widehat{\nabla}_{\phi_t}\mathcal{L}_{t+1}
=
g_{t+1}^{\top}
\frac{\partial \Delta \theta_t(\phi_t)}
{\partial \phi_t}.
\end{equation}

The policy coefficients are then updated by
\begin{equation}
\phi_{t+1}
=
\phi_t
-
\eta_\phi
\widehat{\nabla}_{\phi_t}\mathcal{L}_{t+1},
\end{equation}
where \(\eta_\phi\) denotes the policy learning rate.

This update provides a truncated online estimate of how the previous policy choice affected the subsequent loss. The moment estimates, variance estimates, and gradient-agreement signal used to compute \(\Delta \theta_t\) are treated as fixed with respect to \(\phi_t\) in this one-step policy update. Thus, the method avoids offline optimizer search, a separate meta-training phase, and backpropagation through the full optimization trajectory.

Because the policy contains only \(3(d+1)\) scalar coefficients, the additional policy-state overhead is small. 

\subsection{Experimental Setup}

The experiments evaluate the proposed optimizer, \textsc{PILOT}, on supervised image-classification benchmarks using two datasets and two neural architectures. The evaluation uses FashionMNIST~\cite{xiao2017fashion} as a grayscale clothing-image benchmark and CIFAR-10~\cite{krizhevsky2009learning} as a natural-image benchmark with higher visual variability. The experiments include two model families: a standard convolutional neural network (CNN) and ResNet-18~\cite{he2016deep}, which represents a deeper residual convolutional architecture. The optimizer comparison includes Adam~\cite{kingma2015adam}, AdamW~\cite{loshchilov2019decoupled}, Lion~\cite{chen2023symbolic}, Sophia~\cite{liu2023sophia}, AdaBelief~\cite{zhuang2020adabelief}, and \textsc{PILOT}. The experimental design evaluates each optimizer on FashionMNIST and CIFAR-10 using both CNN and ResNet-18, allowing comparison across simpler and more difficult datasets as well as standard and deeper convolutional architectures.

\subsection{Evaluation}

The evaluation uses four main metrics: classification accuracy, validation loss, loss variance, and mean gradient norm. Classification accuracy measures final predictive performance, with higher values indicating better classification results. Validation loss measures the held-out objective value, with lower values indicating better convergence and generalization behavior. Loss variance measures the stability of the training trajectory by quantifying loss variability across training epochs; lower loss variance indicates smoother optimization. Mean gradient norm measures the average magnitude of parameter gradients during training, providing an additional indicator of optimization stability and gradient behavior. Together, these metrics characterize final performance, convergence quality, training stability, and update dynamics.

\section{Results}
This section reports the empirical performance of \textsc{PILOT} across FashionMNIST and CIFAR-10 using CNN and ResNet-18 architectures. The results compare \textsc{PILOT} against several adaptive optimization baselines in terms of final validation accuracy, validation loss, and convergence stability. The section then examines the contribution of individual policy components through ablation studies.
\begin{figure}[t]
\centering
\includegraphics[width=\linewidth]{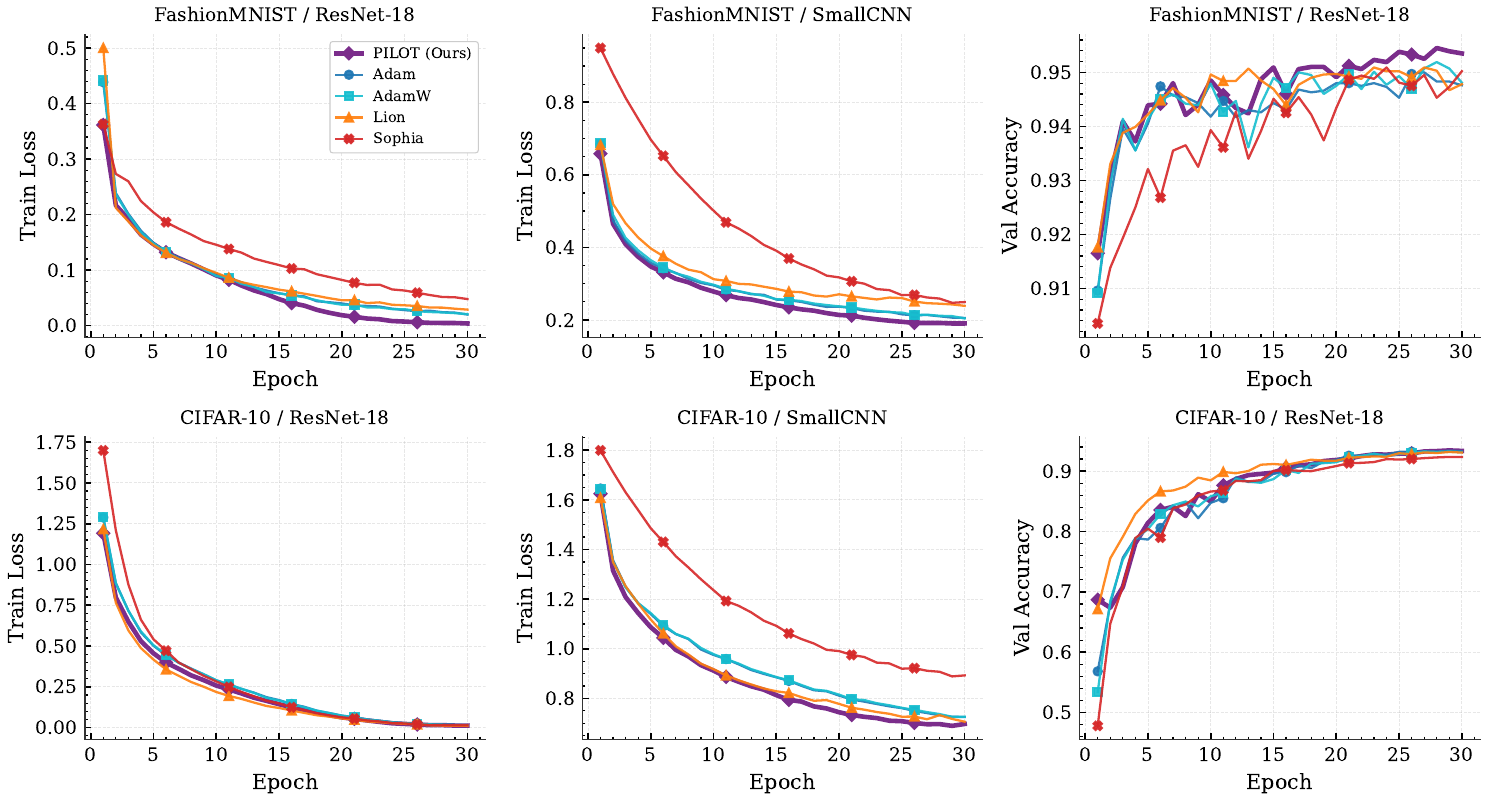}
\caption{Training loss (left, middle) and validation accuracy (right) across FashionMNIST (top) and CIFAR-10 (bottom) over 30 epochs.}
\label{fig:loss_accuracy}
\end{figure}

\subsection{CNN Results Across Datasets}

For the CNN architecture, PILOT achieves the highest accuracy on both datasets. On FashionMNIST, PILOT reaches 94.13\% accuracy, outperforming Adam (93.28\%), AdamW (93.22\%), and all other baselines. PILOT also obtains the lowest validation loss of 0.1719. On CIFAR-10, PILOT achieves 81.94\% accuracy, surpassing Lion (80.87\%), Adam (79.91\%), and AdamW (79.74\%), while also obtaining the best validation loss (0.5302) and the lowest loss variance (0.0073). The complete CNN results are reported in Table~\ref{tab:cnn_results} and Fig \ref{fig:loss_accuracy}.

\begin{table}[H]
\centering
\caption{CNN benchmark results on FashionMNIST and CIFAR-10. Accuracy is reported as percentage. Best values within each dataset group are shown in bold.}
\label{tab:cnn_results}
\begin{tabular}{llccc}
\toprule
Dataset & Optimizer & Accuracy (\%) $\uparrow$ & Val Loss $\downarrow$ & Loss Var. $\downarrow$ \\
\midrule
\multirow{6}{*}{FashionMNIST}
& Adam & 93.28 & 0.1957 & \textbf{0.0033} \\
& AdamW & 93.22 & 0.1944 & 0.0034 \\
& Lion & 92.91 & 0.2091 & 0.0041 \\
& Sophia & 89.14 & 0.3136 & 0.0122 \\
& AdaBelief & 93.66 & 0.1822 & 0.0046 \\
& PILOT (Ours) & \textbf{94.13} & \textbf{0.1719} & 0.0045 \\
\midrule
\multirow{5}{*}{CIFAR-10}
& Adam & 79.91 & 0.5794 & 0.0103 \\
& AdamW & 79.74 & 0.5870 & 0.0102 \\
& Lion & 80.87 & 0.5487 & 0.0105 \\
& Sophia & 76.46 & 0.6874 & 0.0099 \\
& PILOT (Ours) & \textbf{81.94} & \textbf{0.5302} & \textbf{0.0073} \\
\bottomrule
\end{tabular}
\end{table}

\subsection{ResNet-18 Results Across Datasets}

For the deeper ResNet-18 architecture, PILOT achieves the highest accuracy on both datasets. On FashionMNIST, PILOT reaches 95.71\% accuracy, outperforming AdaBelief (95.33\%), AdamW (95.19\%), and all other baselines. Sophia obtains the lowest validation loss of 0.1707. On CIFAR-10, PILOT achieves 93.42\% accuracy, outperforming Adam (93.18\%), AdamW (92.90\%), and Lion (92.71\%). Adam achieves the lowest validation loss (0.2140), while PILOT obtains the lowest loss variance (0.0001), indicating exceptionally stable convergence. The complete ResNet-18 results are reported in Table~\ref{tab:resnet_results}  and Fig \ref{fig:loss_accuracy}.
\begin{table}[h]
\centering
\caption{ResNet-18 benchmark results on FashionMNIST and CIFAR-10. Accuracy is reported as percentage. Best values within each dataset group are shown in bold.}
\label{tab:resnet_results}
\begin{tabular}{llccc}
\toprule
Dataset & Optimizer & Accuracy (\%) $\uparrow$ & Val Loss $\downarrow$ & Loss Var. $\downarrow$ \\
\midrule
\multirow{6}{*}{FashionMNIST}
& Adam & 95.00 & 0.2104 & \textbf{0.0004} \\
& AdamW & 95.19 & 0.2077 & \textbf{0.0004} \\
& Lion & 95.09 & 0.2034 & 0.0007 \\
& Sophia & 95.09 & \textbf{0.1707} & 0.0010 \\
& AdaBelief & 95.33 & 0.1711 & 0.0056 \\
& PILOT (Ours) & \textbf{95.71} & 0.2690 & 0.0030 \\
\midrule
\multirow{5}{*}{CIFAR-10}
& Adam & 93.18 & \textbf{0.2140} & 0.0073 \\
& AdamW & 92.90 & 0.2514 & 0.0066 \\
& Lion & 92.71 & 0.2908 & 0.0067 \\
& Sophia & 91.76 & 0.3223 & 0.0068 \\
& PILOT (Ours) & \textbf{93.42} & 0.2496 & \textbf{0.0001} \\
\bottomrule
\end{tabular}
\end{table}
\subsection{Optimization Stability Analysis}
\begin{figure}[h]
\centering
\includegraphics[width=\linewidth]{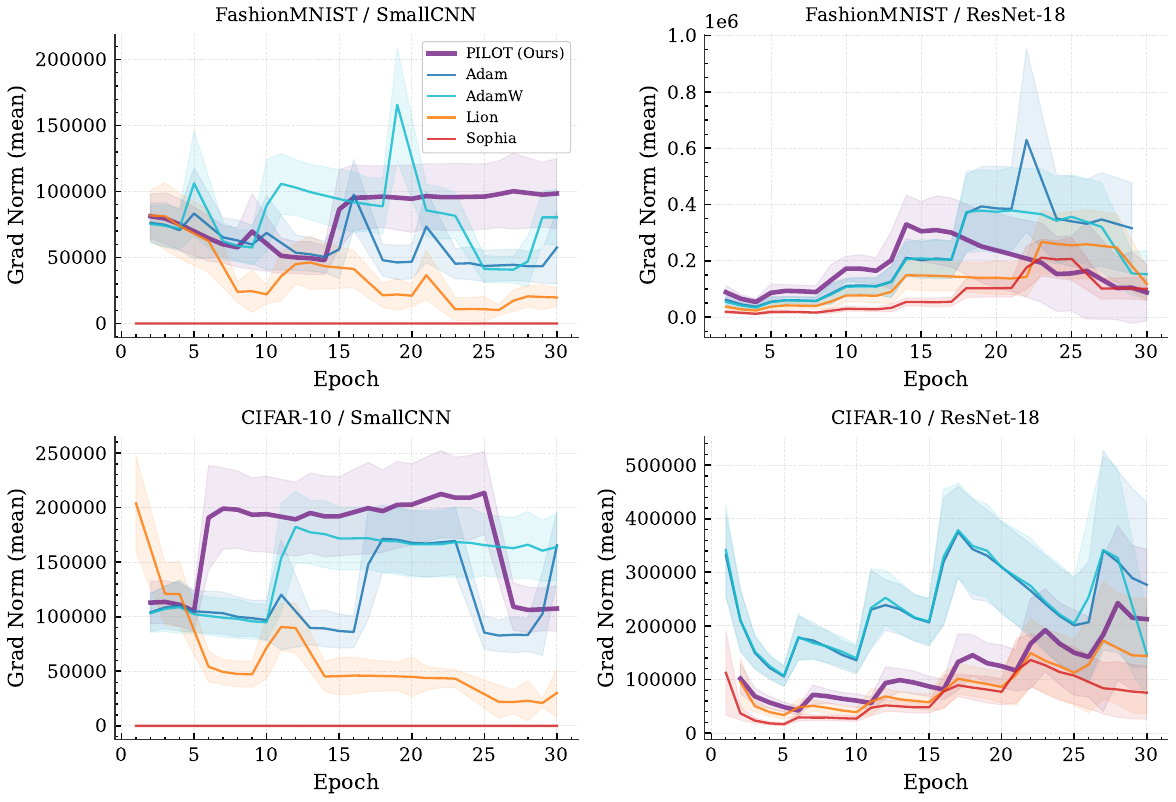}
\caption{Mean gradient norm (with standard-deviation bands) across training epochs for all configurations.}
\label{fig:grad_norm}
\end{figure}

Gradient-norm dynamics provide an additional view of optimizer stability across datasets and architectures, as shown in Figure~\ref{fig:grad_norm}. PILOT generally maintains smoother and more controlled gradient behavior than competing methods, particularly in the ResNet-18 experiments. Adam and AdamW exhibit larger gradient spikes and higher variability, especially on CIFAR-10, while Lion and Sophia often produce lower-magnitude but less adaptive update behavior. In contrast, PILOT maintains relatively stable gradient trajectories while preserving strong classification performance, suggesting that policy-conditioned updates improve optimization stability without overly suppressing gradient dynamics.

\subsection{Ablation Study}

\subsubsection{Effect of Momentum Reliance}

This ablation evaluates the role of momentum reliance by fixing \(p_{m,t}=1\), which forces the update rule to rely fully on the momentum estimate instead of learning the interpolation between the momentum estimate and the current gradient. Compared with the fully learned \textsc{PILOT} variant, this fixed-momentum setting generally reduces performance. On CIFAR-10 with CNN, accuracy decreases from \(80.11\%\) to \(78.97\%\), corresponding to a drop of \(1.14\) percentage points, while validation loss increases from \(0.5757\) to \(0.6159\). This represents the largest degradation observed for this ablation, suggesting that adaptive control of momentum reliance plays an important role in the more challenging CIFAR-10 setting.

A similar but smaller degradation appears on CIFAR-10 with ResNet-18, where accuracy decreases from \(93.26\%\) to \(93.11\%\). On FashionMNIST with ResNet-18, fixing \(p_{m,t}=1\) also reduces accuracy from \(95.71\%\) to \(95.40\%\). These results indicate that learning \(p_{m,t}\) helps the optimizer adjust its reliance on accumulated momentum according to recent gradient behavior. This adaptivity can be useful when gradients become noisy or inconsistent, since excessive reliance on historical momentum may produce less suitable update directions.

The main exception occurs on FashionMNIST with CNN, where fixing \(p_{m,t}=1\) improves accuracy from \(91.99\%\) to \(93.16\%\) and reduces validation loss from \(0.2357\) to \(0.1975\). This result suggests that a fixed momentum-dominated update can be sufficient in simpler low-capacity settings, where the optimization trajectory may not require strong policy-driven adjustment. Overall, the ablation shows that momentum reliance contributes most clearly in the more difficult classification settings, while fixed momentum can remain competitive in simpler CNN configurations.

\subsubsection{Effect of Variance-Normalization Strength}

This ablation evaluates the role of variance-normalization strength by fixing \(p_{v,t}=0.5\), which forces the update rule to use Adam-style square-root scaling instead of learning the normalization exponent. The fixed-normalization setting produces mixed behavior across datasets and architectures. On CIFAR-10, the fully learned \textsc{PILOT} variant remains stronger. With CNN, accuracy decreases from \(80.11\%\) to \(79.69\%\) when \(p_{v,t}\) is fixed, while validation loss increases from \(0.5757\) to \(0.5833\). With ResNet-18, accuracy decreases from \(93.26\%\) to \(93.05\%\), while validation loss remains nearly unchanged. These results suggest that learning \(p_{v,t}\) provides a small but consistent advantage on CIFAR-10 by allowing the optimizer to adjust variance-normalization strength instead of always applying full Adam-style scaling.

The opposite behavior appears on FashionMNIST with CNN. In this setting, fixing \(p_{v,t}=0.5\) gives the strongest result among the ablation variants, increasing accuracy from \(91.99\%\) to \(93.44\%\) and reducing validation loss from \(0.2357\) to \(0.1940\). This result suggests that full variance normalization can be effective in simpler CNN settings, where the additional flexibility of a learned normalization exponent may not be necessary. For FashionMNIST with ResNet-18, the fully learned variant remains slightly stronger in accuracy, achieving \(95.71\%\) compared with \(95.58\%\) for the fixed-\(p_{v,t}\) variant.

Overall, the \(p_{v,t}\) ablation shows that variance-normalization strength depends on dataset complexity and model capacity. Fixed Adam-style normalization can remain sufficient, or even beneficial, in simpler settings, while learning \(p_{v,t}\) provides clearer benefits on CIFAR-10, where the optimizer benefits from dynamically adjusting the amount of adaptive scaling.

\subsubsection{Effect of Sign-Based Policy}

Fixing $p_s=0$ often improves stability-related metrics but does not consistently improve final accuracy. On CIFAR-10 with CNN, accuracy decreases from $80.11\%$ to $79.58\%$ and validation loss increases from $0.5757$ to $0.5897$. However, loss variance decreases from $0.04516$ to $0.03783$, indicating a smoother training trajectory. This suggests that removing sign-based behavior can improve stability, but the resulting smoother optimization path does not necessarily lead to better predictive performance.

A similar trade-off appears in the ResNet-18 experiments. On CIFAR-10 with ResNet-18, fixing $p_s=0$ decreases accuracy from $93.26\%$ to $93.06\%$, although it improves validation loss from $0.2491$ to $0.2347$. On FashionMNIST with ResNet-18, accuracy decreases from $95.71\%$ to $95.24\%$, while validation loss improves from $0.2040$ to $0.1919$. These results suggest that disabling sign-based behavior can sometimes improve loss minimization, but this improvement does not consistently translate into higher classification accuracy.

On FashionMNIST with CNN, fixing $p_s=0$ improves accuracy from $91.99\%$ to $92.79\%$ and reduces validation loss from $0.2357$ to $0.2020$. However, this variant still remains below the best result in that setting, which is obtained by fixing $p_v=0.5$. Overall, the $p_s$ ablation indicates that sign-based behavior plays an important role in the stability--accuracy trade-off. Fixing $p_s=0$ can produce smoother or lower-loss training in some cases, but the fully learned variant usually achieves better accuracy, particularly in the ResNet-18 settings. This suggests that adaptive sign behavior helps the optimizer benefit from magnitude compression when useful, rather than being restricted to purely magnitude-sensitive updates.

\subsection{Policy Transferability}

A natural question is whether the learned policy captures optimization dynamics that generalize across datasets, or whether the polynomial coefficients $\phi$ are specific to the task on which they were learned. To test this, a policy is first trained on CIFAR-10 using ResNet-18, then frozen and applied to FashionMNIST with a fresh model initialization. The frozen policy is compared against AdamW and a fresh \textsc{PILOT} instance that learns its own policy on the target dataset.

\begin{figure}[h]
\centering
\includegraphics[width=\linewidth]{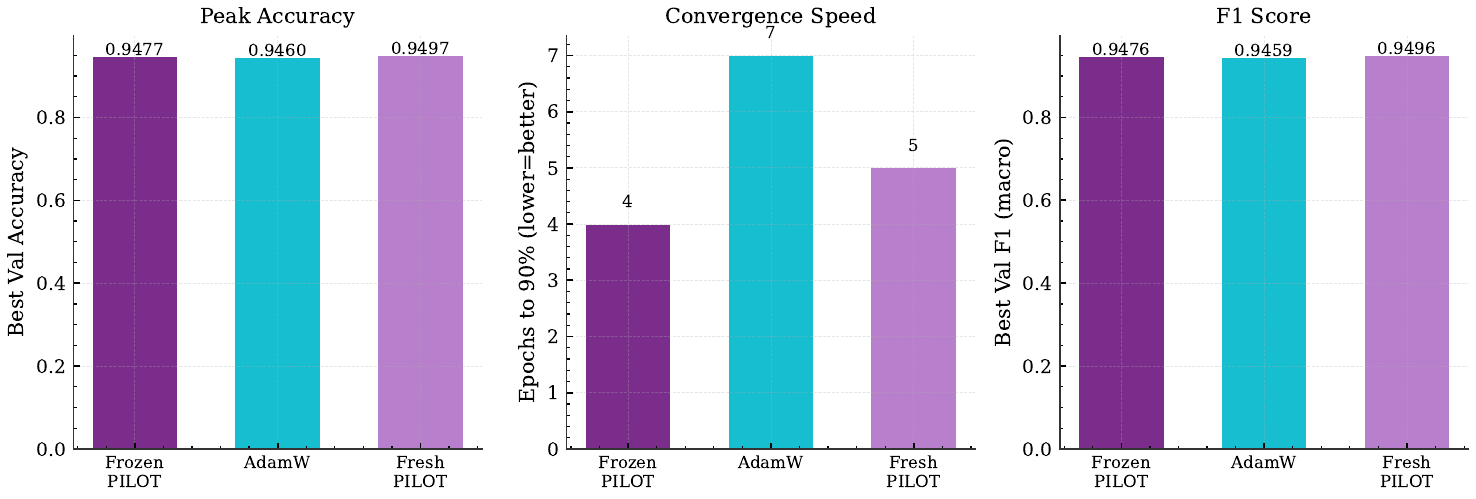}
\caption{Policy-transfer summary on FashionMNIST / ResNet-18. The frozen \textsc{PILOT} policy, learned entirely on CIFAR-10, outperforms AdamW and nearly matches a fresh \textsc{PILOT} trained with active meta-gradients.}
\label{fig:transfer_summary}
\end{figure}

Figure~\ref{fig:transfer_summary} summarizes the results. The frozen policy outperforms AdamW in accuracy, F1 score, and convergence speed, reaching 90\% validation accuracy in 4~epochs compared with 7 for AdamW. The fresh \textsc{PILOT} variant achieves the strongest final accuracy (94.97\%) through additional target-specific adaptation, but the gap relative to the frozen variant is small (0.20 percentage points). This suggests that most of the policy's benefit is already captured by the source-task coefficients, and that the learned polynomial encodes broadly applicable optimization dynamics rather than dataset-specific patterns.

\section{Conclusion}

This study introduced \textsc{PILOT}, an adaptive first-order optimizer that adjusts its update behavior online using gradient-direction agreement as a descriptor of local update consistency. Instead of relying on a fixed update structure throughout training, \textsc{PILOT} learns a compact policy that modulates momentum reliance, variance-normalization strength, and sign-based behavior. This enables the optimizer to adapt between Adam-style, raw-gradient, and sign-dominated update regimes while retaining a practical first-order formulation.

The experimental results show that \textsc{PILOT} achieves strong performance across the evaluated image-classification settings. In particular, the optimizer obtains the highest accuracy across the reported CNN and ResNet-18 experiments on FashionMNIST and CIFAR-10. The ablation study further supports the importance of the learned policy variables, showing that \(p_{m,t}\), \(p_{v,t}\), and \(p_{s,t}\) contribute complementary effects by controlling optimizer memory, variance normalization, and the balance between magnitude-sensitive and sign-based update directions. The policy-transfer experiment provides additional evidence that the learned polynomial captures broadly applicable optimization dynamics: a policy trained on CIFAR-10 transfers to FashionMNIST without further adaptation and outperforms AdamW in both accuracy and convergence speed. These findings suggest that adapting the update structure during training can be beneficial when optimization dynamics vary across different stages of learning, and that the resulting policy generalizes beyond the dataset on which it was learned.

Overall, \textsc{PILOT} demonstrates that a small online policy can provide meaningful adaptivity within a first-order optimizer. The results support the broader view that optimizer behavior does not need to remain fixed throughout training, especially when the loss landscape changes over time. By combining interpretable policy variables with online adaptation, \textsc{PILOT} provides a promising direction for flexible optimizer design in deep neural network training.

\subsection{Limitations}

The current evaluation remains limited to relatively small image-classification benchmarks and convolutional architectures. Although the policy-transfer experiment provides initial evidence that the learned policy generalizes across datasets, the generality of \textsc{PILOT} has not yet been established for larger datasets, transformer-based architectures, sequence models, language models, or dense prediction tasks such as object detection and semantic segmentation. Additional experiments are therefore needed to determine whether the same adaptive behavior and cross-task transferability extend to more complex and large-scale optimization settings.

The current study also focuses mainly on predictive performance and training stability. A more detailed analysis of policy dynamics, learned parameter trajectories, hyperparameter sensitivity, and computational overhead would provide a deeper understanding of how \textsc{PILOT} behaves during training. Although the method avoids offline optimizer search and explicit second-order estimation, the online policy update still introduces additional computation and optimizer-state overhead, which should be quantified more systematically.

\subsection{Future Work}

Future work should evaluate \textsc{PILOT} on larger-scale datasets and more demanding architectures, including Vision Transformers, Swin Transformers, large residual networks, and language models. Further investigation should also analyze how the learned policy variables evolve during training and whether their behavior corresponds to identifiable optimization regimes, such as stable descent, noisy gradients, sharp curvature, or plateau regions.

Another important direction is to extend the policy formulation with additional online signals, such as gradient variance, curvature approximations, update norms, or loss-change history. Finally, theoretical analysis of stability and convergence properties would strengthen the foundation of the proposed method and help clarify when policy-based online adaptation should outperform fixed update rules.

\bibliographystyle{unsrt}

\bibliography{bib}
\clearpage
\appendix

\section{Additional Method Details}

\subsection{\textsc{PILOT} Algorithm}
Algorithm~\ref{alg:pilot} summarizes the training procedure of \textsc{PILOT}. At each iteration, the optimizer computes the stochastic gradient and evaluates the directional agreement between the current and previous gradients. The exponential moving average \(\rho_t\) smooths this agreement signal and provides the input to the polynomial policy. The policy then produces the control variables \(p_{m,t}\), \(p_{v,t}\), and \(p_{s,t}\), which control momentum reliance, variance-normalization strength, and sign-based behavior in the update rule. After the model parameters are updated using the policy-conditioned direction, the policy coefficients \(\phi\) are updated online using the subsequent training signal. This procedure allows \textsc{PILOT} to refine its update behavior during training without offline optimizer search, an auxiliary optimizer network, or a separate meta-training phase.

\begin{algorithm}[ht]
\caption{\textsc{PILOT} Optimizer}
\label{alg:pilot}
\begin{algorithmic}[1]
\REQUIRE polynomial degree \(d\), learning rates \(\eta\), \(\eta_\phi\), smoothing coefficient \(\gamma\)
\STATE Initialize parameters \(\theta_0\), moments \(m_0=0\), \(v_0=0\), agreement state \(\rho_0=0\), and policy coefficients \(\phi \in \mathbb{R}^{3(d+1)}\)
\STATE Initialize previous gradient \(g_0=0\)
\FOR{\(t = 1\) to \(T\)}
    \STATE Compute stochastic gradient \(g_t = \nabla_\theta \mathcal{L}_t(\theta_t)\)
    \IF{\(t=1\)}
        \STATE Set \(r_t = 0\)
    \ELSE
        \STATE Compute gradient-direction agreement \(r_t\)
    \ENDIF
    \STATE Update smoothed agreement signal \(\rho_t = \gamma \rho_{t-1} + (1-\gamma)r_t\)
    \STATE Compute policy variables \(p_{m,t}\), \(p_{v,t}\), and \(p_{s,t}\) from \(\rho_t\)
    \STATE Update moment estimates \(m_t\) and \(v_t\)
    \STATE Compute policy-controlled direction \(n_t\)
    \STATE Update model parameters \(\theta_{t+1}\)
    \STATE Update policy coefficients \(\phi\) using the one-step online policy update
    \STATE Store \(g_t\) as the previous gradient
\ENDFOR
\end{algorithmic}
\end{algorithm}

\section{Experimental Details}
\label{app:details}

\subsection{Training Configuration}

The training protocol uses 30 epochs and cross-entropy loss for all models. Automatic mixed precision (AMP) is enabled across all configurations. The learning rate follows a cosine annealing schedule; for ResNet-18 configurations, a linear warmup phase of 3 epochs precedes cosine decay. All experiments use a batch size of 128. Table~\ref{tab:training_config} reports the training hyperparameters shared across optimizers for each dataset--architecture configuration.

\begin{table}[h]
\centering
\caption{Training hyperparameters shared across optimizers for each configuration.}
\label{tab:training_config}
\begin{tabular}{llcccc}
\toprule
Dataset & Architecture & Learning Rate & Weight Decay & Warmup Epochs & Batch Size \\
\midrule
FashionMNIST & CNN & $1\times10^{-3}$ & $1\times10^{-4}$ & 0 & 128 \\
FashionMNIST & ResNet-18 & $1\times10^{-4}$ & $1\times10^{-2}$ & 3 & 128 \\
CIFAR-10 & CNN & $1\times10^{-3}$ & $1\times10^{-4}$ & 0 & 128 \\
CIFAR-10 & ResNet-18 & $1\times10^{-3}$ & $1\times10^{-4}$ & 3 & 128 \\
\bottomrule
\end{tabular}
\end{table}

\subsection{Optimizer-Specific Hyperparameters}

Adam, AdamW, and \textsc{PILOT} use moment coefficients \(\beta_1=0.9\) and \(\beta_2=0.999\). Lion uses \(\beta_1=0.9\) and \(\beta_2=0.99\), with the learning rate scaled to \(\eta/3\) and weight decay scaled to \(3\lambda\), following the scaling recommendations of Chen et al.~\cite{chen2023symbolic}. Sophia uses \(\beta_1=0.965\), \(\beta_2=0.99\), clipping threshold \(\rho_{\mathrm{Sophia}}=0.04\), learning rate \(2\eta\), and a diagonal Hessian estimate updated every 10 steps.

The \textsc{PILOT}-specific hyperparameters---the smoothing coefficient \(\gamma\), the policy learning rate \(\eta_\phi\), and the polynomial degree \(d\)---are selected independently for each configuration via Bayesian optimization using the Tree-structured Parzen Estimator (TPE) with ASHA early stopping. Each search consists of 30--40 trials trained for 25 epochs. The search ranges are \(\gamma \in [0.85,0.99]\), \(\eta_\phi \in [5\times10^{-4},5\times10^{-2}]\) with log-uniform sampling, and \(d \in \{1,2,3,4\}\). Table~\ref{tab:pilot_hparams} reports the selected values.

\begin{table}[h]
\centering
\caption{Selected \textsc{PILOT} hyperparameters for each configuration.}
\label{tab:pilot_hparams}
\begin{tabular}{llccc}
\toprule
Dataset & Architecture & \(\gamma\) & \(\eta_\phi\) & Degree \(d\) \\
\midrule
CIFAR-10 & CNN & 0.882 & 0.00312 & 1 \\
CIFAR-10 & ResNet-18 & 0.950 & 0.00500 & 2 \\
FashionMNIST & CNN & 0.950 & 0.01000 & 2 \\
FashionMNIST & ResNet-18 & 0.957 & 0.00273 & 3 \\
\bottomrule
\end{tabular}
\end{table}

\subsection{Data Augmentation and Preprocessing}

For the CNN architecture, the preprocessing pipeline applies random cropping with 4-pixel padding on CIFAR-10 and 2-pixel padding on FashionMNIST, together with random horizontal flipping. For ResNet-18, the pipeline resizes inputs to \(224 \times 224\) pixels using random resized crops with scale range \([0.75,1.0]\) during training and center crops during evaluation. FashionMNIST images are replicated to three channels when used with ResNet-18 to match the expected input format. All inputs are normalized using dataset-specific channel means and standard deviations.

\subsection{Reproducibility}
All experiments use a fixed random seed of 42 for weight initialization and data loading. The ResNet-18 model on FashionMNIST uses ImageNet-pretrained initialization; all other models use random initialization. The optimizer implementation and experiment scripts are provided as supplemental material.

\subsection{Compute Resources}
\label{app:compute}

The benchmark experiments reported in Table~\ref{tab:cnn_results} and Table~\ref{tab:resnet_results} ran on an Ibex HPC cluster equipped with NVIDIA Tesla V100-SXM2-32GB GPUs. Hyperparameter tuning ran on Google Colab instances provisioned with NVIDIA A100 GPUs. The final benchmark includes 4 dataset--architecture configurations and the evaluated optimizer baselines, with additional runs for the ablation study. Preliminary experiments and hyperparameter search, consisting of 30--40 trials per configuration trained for 25 epochs each, required additional compute beyond the final reported benchmark results.
\begin{figure}[H]
\centering
\includegraphics[width=0.85\linewidth]{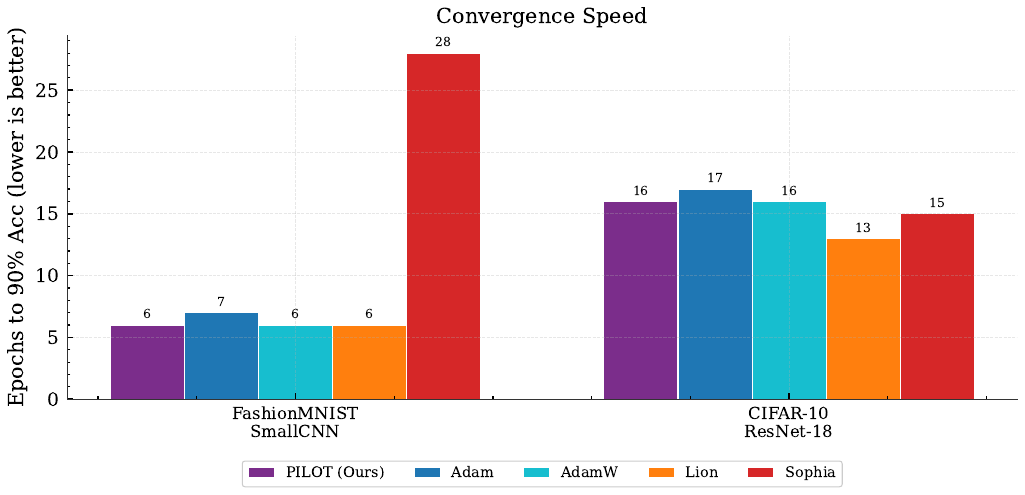}
\caption{Epochs to reach 90\% validation accuracy (lower is better).}
\label{fig:convergence}
\end{figure}

Figure~\ref{fig:convergence} compares convergence speed across optimizers using the number of epochs required to reach 90\% validation accuracy. PILOT reaches the target accuracy competitively across both settings, matching or outperforming most adaptive baselines. On FashionMNIST with SmallCNN, PILOT reaches the target in 6 epochs, while Sophia converges substantially slower. On CIFAR-10 with ResNet-18, Lion converges fastest, although PILOT remains competitive while achieving stronger final accuracy and more stable optimization behavior overall.

\subsection{Policy-Transfer Experiment Details}
\label{app:transfer}

The policy-transfer experiment reported in Section~4.5 proceeds in two phases. In the source phase, \textsc{PILOT} is trained on CIFAR-10 with ResNet-18 for 30 epochs using the hyperparameters in Table~\ref{tab:pilot_hparams} ($\gamma=0.95$, $\eta_\phi=0.005$, degree~$d=2$), producing a set of learned policy coefficients $\phi^* \in \mathbb{R}^{9}$.

In the target phase, a fresh ResNet-18 is initialized from random weights and three identical copies of this initialization are trained on FashionMNIST for 30 epochs with $\text{lr}=10^{-4}$, $\text{wd}=10^{-2}$, and a 3-epoch linear warmup followed by cosine annealing. The three conditions are:
\begin{itemize}
    \item \textbf{Frozen \textsc{PILOT}}: the source-task coefficients $\phi^*$ are injected into the optimizer and the meta-learning rate is set to $\eta_\phi=0$, so the polynomial still maps $\rho_t$ to policy variables at each step but the coefficients are never updated.
    \item \textbf{AdamW}: standard AdamW baseline with the same learning rate and weight decay.
    \item \textbf{Fresh \textsc{PILOT}}: initialized with default policy coefficients and trained with active meta-gradients ($\eta_\phi=0.005$).
\end{itemize}
All three conditions use degree~$d=2$ to ensure equal policy expressiveness and share the same model initialization, data augmentation, and learning-rate schedule. The random seed for the target-phase initialization differs from the source phase to ensure independent weights.

\end{document}